\documentclass[letterpaper, 10 pt, conference]{ieeeconf}  %

\IEEEoverridecommandlockouts                              %

\usepackage{graphicx}
\usepackage{amsmath} %
\usepackage{amssymb}  %
\usepackage{bm}
\usepackage{siunitx}
\usepackage{caption}
\usepackage{subcaption}
\usepackage{cite}  %
\usepackage{algorithm}
\usepackage[noend]{algpseudocode}
\newcommand*\Let[2]{\State #1 $\gets$ #2}
\algrenewcommand\alglinenumber[1]{
    {\sf\footnotesize\addfontfeatures{Colour=888888,Numbers=Monospaced}#1}}
\algrenewcommand\algorithmicrequire{\textbf{Precondition:}}
\algrenewcommand\algorithmicensure{\textbf{Postcondition:}}

\graphicspath{{figs/}}

\title{\LARGE \bf
DIJE: Dense Image Jacobian Estimation\\ for Robust Robotic Self-Recognition and Visual Servoing
}

\author{Yasunori Toshimitsu$^{1}$, Kento Kawaharazuka$^{1}$, Akihiro Miki$^{1}$, Kei Okada$^{1}$, and Masayuki Inaba$^{1}$
\thanks{$^{1}$The authors are with the Department of Mechano-Informatics, Graduate School of Information Science and Technology, The University of Tokyo, 7-3-1 Hongo, Bunkyo-ku, Tokyo, 113-8656, Japan.
        {\tt\small [toshimitsu, kawaharazuka, miki, k-okada, inaba]@jsk.t.u-tokyo.ac.jp}
        }
}

\begin{document}

\maketitle
\thispagestyle{empty}
\pagestyle{empty}

\begin{abstract}
For robots to move in the real world, they must first correctly understand the state of its own body and the tools that it holds.
In this research, we propose DIJE, an algorithm to estimate the image Jacobian for every pixel. It is based on an optical flow calculation and a simplified Kalman Filter that can be efficiently run on the whole image in real time. It does not rely on markers nor knowledge of the robotic structure. We use the DIJE in a self-recognition process which can robustly distinguish between movement by the robot and by external entities, even when the motion overlaps. We also propose a visual servoing controller based on DIJE, which can learn to control the robot's body to conduct reaching movements or bimanual tool-tip control. The proposed algorithms were implemented on a physical musculoskeletal robot and its performance was verified. We believe that such global estimation of the visuomotor policy has the potential to be extended into a more general framework for manipulation.
\end{abstract}

\section{Introduction}
While robots have predominantly been used in factory environments for decades, they have only recently started to be used in everyday environments, which can be unstructured and unpredictable. These uncertainties exist both in the external environment and the robot's own body. In particular, we consider that it is important for the robot to be able to discover its own body while actively moving within the environment.

We can consider the problem of visual self-body control as a two component problem. For one, the robot must be able to recognize the \textit{extent of its own body}, i.e. a self-recognition problem. It must be generalize to recognize novel visual features such as new tools in its hands, while ignoring external movements not caused by the robot. Also, the robot must recognize \textit{how to control its own body}. This is akin to acquiring an internal model of the robot movement usable in control, a simple example being the manipulator Jacobian matrix.

To achieve this, a holistic understanding of the visuomotor system is required, rather than calculating values discretely for predetermined keypoints or markers along the robot. To that end, we propose DIJE (Dense Image Jacobian Estimator), a method to densely (i.e. for every pixel of the image) estimate the image Jacobian based on exteroceptive camera images and proprioceptive joint sensor data. It does not require \textit{a priori} knowledge of the robot's kinematic structure, and the resulting dense image Jacobian encodes the global relationship between visual features and joint information for the whole image. 

Multiple image Jacobian estimation methods have been proposed \cite{Hosoda1994-ng, Mao2012-jn, Lv2006-xs, Farahmand2007-iw, Qian2002-zz}, but to our knowledge, this work is the first to apply it densely on the image. To enable this, we propose a simplified Kalman Filter-based estimation algorithm for the image Jacobian, and an update rule to shift the image Jacobian estimation value in each timestep based on the robot movement.
DIJE enables a unified approach to self-recognition and visual servoing control.
We propose using the output of DIJE to a self-recognition algorithm which can robustly predict the extent of the robot's own body even in the presence of external movements, and to a markerless visual servoing controller which can learn to control the robot's body and the tools in its hands.

The contributions of this work are as follows:
\begin{itemize}
        \item a dense estimation method for the image Jacobian implemented by a recursive algorithm, based on the Kalman Filter and a dense update method across timesteps, robust to non robot-induced external movements
        \item a dense binary labeling algorithm of the robot's self body based on the output of DIJE, capable of distinguishing between self-induced and external movement 
        \item a markerless visual servoing controller based on the output of DIJE
\end{itemize}
All of the algorithms presented require no prior knowledge of the robot structure, and can be run in real time on a conventional notebook PC.

\section{Related Work}
\subsection{Robotic Self Recognition}
In a robotic self recognition system, the robot attempts to locate and distinguish its own body from the environment in the visual (or depth) image. They can be split into methods requiring prior knowledge of the robotic structure \cite{Florence2020-zg, Lambrecht2019-ms, Rauch2018-vu,Widmaier2016-qv}, and those that don't \cite{Metta2004-kw,Michel2004-yn,Kemp2006-xw}.

For methods requiring \textit{a priori} knowledge, the recognition algorithm is trained based on datasets synthesized from a simulation of the robot. Machine learning algorithms such as Mask R-CNN \cite{Florence2020-zg}, CNN \cite{Lambrecht2019-ms}, or random forest \cite{Rauch2018-vu,Widmaier2016-qv} has been used for the recognizer.

In the simulator, perfect knowledge of the robot state is available. 
These recognition models can take advantage of that fact to estimate other features from the image and not just the location or binary mask of the robot body, such as joint angle pose \cite{Lambrecht2019-ms, Rauch2018-vu,Widmaier2016-qv}. However, since a training process is required, the application of these methods to recognition of novel objects attached to the robot (such as tools) is not possible.

Model-less methods for self-recognition usually utilize the motion of the robot to segment its body from the background. 
In the work by Metta and Fitzpatrick, the optical flow is used to detect motion, and as its direction changes depending on joint movement, it is compared to the joint angle measurement to detect the location of the arm \cite{Metta2004-kw}.
Other methods similarly use the detected motion and correlate it with the joint movement, by calculating the mutual information \cite{Kemp2006-xw} or comparing the temporal correlation \cite{Michel2004-yn} to determine whether to attribute it to the robot.

\subsection{Uncalibrated Visual Servoing}
In uncalibrated visual servoing, the manipulator model and camera parameters are not known \textit{a priori} and must be learned online while the robot moves.
The image Jacobian $J$ describes the differential relationship between a point in the image of the robot's body and the robot's joint pose.
\begin{equation}
        \label{eq:image_jacobian}
        \dot{\bm p} = \bm u = J(\bm q) \dot{\bm q}
\end{equation}
where $\bm p$ is the time-varying image coordinate of a point on the robot, $\bm u$ is its time derivative, i.e. the optical flow, and $\dot{\bm q}$ is the joint velocity.
As the pseudoinverse of the image Jacobian can be used to control the image coordinates of the robot, the online estimation of $J$ has been intensely researched for uncalibrated visual servoing, with methods being proposed based on Broyden updating \cite{Hosoda1994-ng}, Kalman Filtering \cite{Qian2002-zz, Lv2006-xs}, support vector regression \cite{Mao2012-jn}, or least-squares fitting \cite{Farahmand2007-iw}. In these works, the estimated image Jacobian has been successfully applied to visual servoing to control robots whose kinematic structures are unknown.

However, many of the uncalibrated visual servoing methods assume that the point of interest being controlled is known \textit{a priori}, for example by markers \cite{Hosoda1994-ng,Qian2002-zz} or is only verified on a simulated model \cite{Farahmand2007-iw, Lv2006-xs,Mao2012-jn} where perfect knowledge is available for where each point of the robot is located in the camera image.

\section{DIJE: Dense Image Jacobian Estimation}
\begin{figure*}
    \centering
    \includegraphics[width=2\columnwidth]{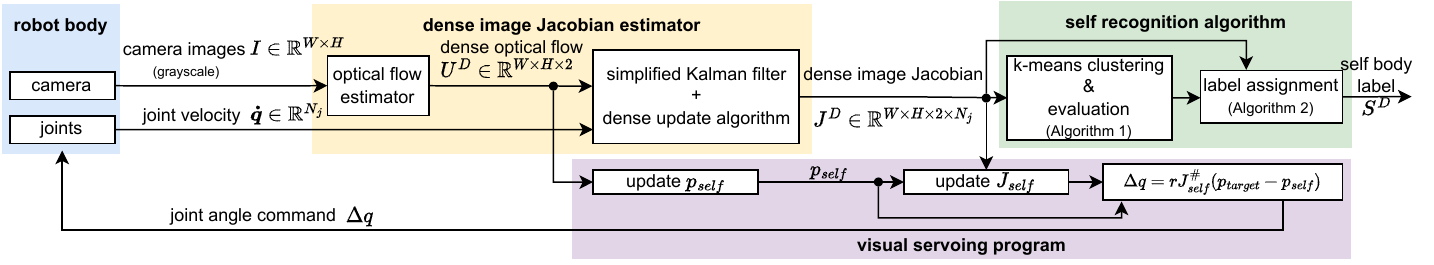}
    \caption{\label{fig:dense_jacobian_estimation}System diagram of DIJE. It combines exteroceptive data (camera images) and proprioceptive data (joint states) to compute the image Jacobian for every pixel.}
\end{figure*}
\begin{figure*}
    \centering
    \includegraphics[width=2\columnwidth]{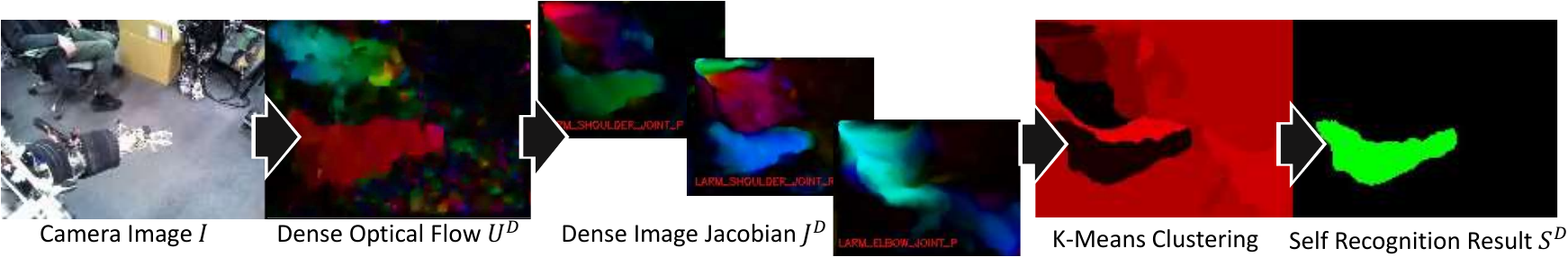}
    \caption{\label{fig:dije_process}Visualization of each processing step in DIJE and the dense self recognition algorithm. $U^D$ and each column of $J^D$ are visualized with the hue and brightness in HSV color space. Each cluster is visualized with different shades of red. Even when there is movement in the background (person moving in a chair) the self recognition algorithm can correctly identify which part of the image belongs to the robot.}
\end{figure*}
In this section, we describe the algorithms used in the estimation of the dense image Jacobian. An overview of the process is shown in Fig~\ref{fig:dense_jacobian_estimation}. In Section~\ref{sec:kf} we describe the Kalman Filter-based estimator that is run on every pixel that estimates the image Jacobian. Then in Section~\ref{sec:update}, the update algorithm is introduced, which is critical to apply the estimation densely across the entire image.

\subsection{Image Jacobian Estimation Formulated as a memory-efficient Kalman Filter}
\label{sec:kf}
Here, the Kalman filter formulation for the image Jacobian estimation is described. This is run separately for each pixel, and thus the computation can be parallelized.
The linear relationship between the optical flow $\bm u$, image Jacobian $J$, and joint velocity $\dot{\bm q}$ shown in \eqref{eq:image_jacobian} is exploited to formulate it as a Kalman filter (KF). The state estimated by the KF is the image Jacobian matrix concatenated into a vector $\bm j$, as
\begin{equation}
\begin{alignedat}{2}
\label{eq:jacobian_to_vector}
        J &= \begin{bmatrix} \bm j_x & \bm j_y \end{bmatrix}^T &&\in \mathbb{R}^{2\times N_j} \\
        \bm j &:= \begin{bmatrix} \bm j_x \\ \bm j_y \end{bmatrix} &&\in \mathbb{R}^{2 N_j}
\end{alignedat}
\end{equation}
where $N_j$ is the number of joints.
Then, by using the optical flow $\bm u \in \mathbb R^2$ as the observation, the observation model can be formulated using an observation matrix $M_q$ which is created from the joint velocity data:
\begin{equation}
\label{eq:mq_definition}
\begin{split}
        \bm u &= M_q \bm j \\
        M_q &:= 
        \begin{bmatrix}
        \dot{\bm q}^T & \\
        & \dot{\bm q}^T
        \end{bmatrix}
\end{split}
\end{equation}
Thus, the model for the KF can be formulated as follows:
\begin{equation}
\label{eq:kalman_filter_jacobian}
\begin{alignedat}{2}
        \bm j_{k} &= update(\bm j_{k-1}) + \bm w_{k} && \text{: state model} \\
        \bm u_{k} &= M_q \bm j_{k} + \bm v_{k} && \text{: observation model}
\end{alignedat}
\end{equation}
Here, the subscript $k$ refers to data for timestep $k$. $\bm w_{k}$ is the process noise which follows a zero mean Gaussian distribution with covariance $Q$ and $\bm v_{k}$ is the observation noise which follows a zero mean Gaussian distribution with covariance $R$. The estimated state $\bm j_{k}$ has covariance $P_{k}$.
$update(\bm j_{k-1})$ is the result of the state transition model for the dense image Jacobian, which updates the value of the estimated image Jacobian of each pixel based on the robot's movement, whose process is described in Section~\ref{sec:update}.
We note that if we were to do a conventional sparse estimation of the image Jacobian, the previous value itself $\bm j_{k-1}$ can be used instead of $update(\bm j_{k-1})$. This is because the point at which the image Jacobian is estimated follows the robot's movement and always represents the same point on the robot, and thus the image Jacobian can be considered to be constant over time.

So far, the formulation of the KF to estimate the image Jacobian is the same as what has been proposed by Qian and Su in 2002 \cite{Qian2002-zz}, apart from the introduction of $update(\bm j_{k-1})$.
However, in this research, the KF must be run on every pixel. In order to make the calculation tractable, we propose to approximate the state covariance matrix $P_k$ as a diagonal matrix, and therefore calculate only the diagonal elements.
In other words, only the variance is considered in the model, and the covariance is disregarded.
Without this assumption, the memory size required to save the covariance matrix becomes $O(N_j^2)$, which becomes intractable as the number of joints $N_j$ increases\footnote{for example, saving the covariance matrix for each pixel in a $320 \times 240$ image for $N_j=5$ with 8-byte floats already amounts to 61MB of data.}. By calculating only the diagonal elements of the covariance matrix, the memory requirement becomes $O(N_j)$. This approximation is important to ensure the scalability of DIJE to multi-joint structures.

Thus, the covariance matrix of the image Jacobian vector $\bm j_k$, $P_k$, can be written as:
\begin{equation}
        P_k = \text{diag}(p_{k,1}, p_{k,2}, \dots, p_{k,2N_j})
\end{equation}
We make some further assumptions; we assume that the first $N_j$ elements respectively match the last $N_j$ elements, i.e.:
\begin{equation}
\label{eq:p_equality}
\begin{split}
        p_{k, 1} &= p_{k, 1 + N_j}\\
        p_{k, 2} &= p_{k, 2 + N_j}\\
        \vdots\\
        p_{k, N_j} &= p_{k, 2N_j}
\end{split}
\end{equation}
This indicates that for each joint, the variance for the horizontal and vertical direction is the same. Then, $P_k$ can be written with a vector $\bm p_k \in \mathbb R^{N_j}$, as
\begin{equation}
        P_k = \text{diag}(\begin{bmatrix}\bm p_k \\ \bm p_k \end{bmatrix})
\end{equation}
By running this formulationn for $P_k$ through the Kalman Filter calculations, it can shown though mathematical induction that \eqref{eq:p_equality} holds for every step $k$, as long as it is true for the initial value.

Using the formulations and assumptions introduced above, the actual calculation of the KF can be formulated as follows.
First, in the prediction step, the \textit{a priori} image Jacobian estimate and its variance is calculated.
\begin{equation}
\label{eq:kalman_filter_state_prediction}
        J_{k|k-1} = update(J_{k-1})
\end{equation}
\begin{equation}
\label{eq:kalman_filter_state_covariance_prediction}
        \bm p_{k|k-1} = \bm p_{k-1} + q \bm 1
\end{equation}
$\bm 1$ is a vector whose elements are all 1.
$q$ is the variance of the process noise, and is assumed to be the same for all elements of the state vector. The process for $update(J_{k-1})$ is described in Section~\ref{sec:update}. Next, in the update step, the \textit{a priori} estimate $J_{k|k-1}$ and variance $\bm p_{k|k-1}$, the joint velocity $\bm{\dot q}$, and the observed optical flow $\bm u_{k}$ is used to calculate the \textit{a posteriori} estimate. The full derivation of these equations is omitted here for the sake of space.
\begin{equation}
\label{eq:kalman_filter_state_update}
        J_{k} = J_{k|k-1} + \frac{(\bm u_{k} - J_{k|k-1} \bm{\dot q}) (\bm p_{k|k-1} \odot \bm{\dot q})^T}{\bm p_{k|k-1} ^T (\bm{\dot q} \odot \bm{\dot q}) + r}
\end{equation}
\begin{equation}
\label{eq:kalman_filter_covariance_update}
        \bm p_{k} = \bm p_{k|k-1} \odot (\bm 1 - \frac{\bm p_{k|k-1} \odot \bm{\dot q}^2}{r + \bm p_{k|k-1}^T\bm{\dot q}^2})
\end{equation}
Here, $r$ is the variance of the observation noise.

We note that this simplified KF-based estimation of the image Jacobian yields an update rule similar to the one proposed by Hosoda et al. in 1994 (some of the notation has been modified to match that of this paper) \cite{Hosoda1994-ng}:
\begin{equation}
\label{eq:hosoda_et_al}
        \Delta J = \frac{(\bm u - J \bm{\dot q}) \bm{\dot q}^T W}{\rho + \bm{\dot q}^T W \bm{\dot q}}
\end{equation}
We can see this as a special case of \eqref{eq:kalman_filter_state_update}, where the observation noise variance $r$ corresponds to the forgetting factor $\rho$, and where the state estimate variance $\bm p$ has a fixed value across every step and corresponds to the weighting matrix $W$.

\subsection{Updating the dense image Jacobian estimate across timesteps}
\label{sec:update}
In this section, we will detail the algorithm to obtain $J_{k|k-1}$ used in \eqref{eq:kalman_filter_state_prediction}, which updates the dense image Jacobian across timesteps. The superscript $D$ will be used to indicate a dense value that is calculated for every pixel. Here, the ideal update method would be to set the same value of the image Jacobian at step $k$ to that of the same point on the robot at step $k-1$.
Thus, we can first consider using the observed optical flow to update the dense image Jacobian, since they describe the movement caused by the robot. This can be done using grid interpolation for each pixel $i$, as
\begin{equation}
\label{eq:dij_update_naive}
        J_{k|k-1, i} \leftarrow \textrm{interpolate}(J_{k-1}^D, \bm x_i - \bm u_{k, i})
\end{equation}
where $\bm x_i$ is the image coordinates of pixel $i$, and thus $\bm x_i - \bm u_{k, i}$ is the location of the point appearing at $\bm x_i$ at the previous timestep.
The $\textrm{interpolate}(\cdot)$ function interpolates on a regular grid, and thus efficient bilinear interpolation functions such as that implemented in scipy.interpolate.interpn\footnote{https://scipy.org/} can be used to calculate this interpolation for all pixels in real time.

However, this method has the issue that movement in the environment may be picked up to falsely update the dense image Jacobian, despite it not being caused by the robot. The optical flow caused by such background movement causes the Jacobian value estimated for the robot's body to "leak out" into the environment.

Thus, we propose to instead use the predicted optical flow calculated from the current estimate of the image Jacobian, to update the dense image Jacobian across timesteps. Thus, the update method for each pixel $i$ is as follows:
\begin{equation}
\label{eq:dij_update}
        J_{k|k-1, i} \leftarrow \textrm{interpolate}(J_{k-1}^D, \bm x_i - J_{k-1, i} \bm{\dot q})
\end{equation}
As long as the image Jacobian is appropriately estimated, the predicted optical flow $J_{k-1, i} \bm{\dot q}$ is not affected by movements not caused by the robot. The effect of this proposed update method is verified in the experiments section.

\section{Application of DIJE to recognition and control}

\subsection{Dense labeling for self-recognition}
\label{sec:self_body_recognition}

In this section, we propose an algorithm for dense binary labeling of the robot's own body, based on the k-means clustering algorithm using the output of DIJE.
Algorithms \ref{alg:self_body_kmeans} and \ref{alg:self_body_label} describe the process.
Algorithm \ref{alg:self_body_kmeans} does k-means clustering and calculates the likeliness that each cluster belongs to the robot's own body, and Algorithm \ref{alg:self_body_label} applies the clustering result to the image to get the dense self-recognition label $S^D$.
The two algorithms are run separately because the k-means calculation in Algorithm \ref{alg:self_body_kmeans} cannot be run in real time, while the dense label $S^D$ must be generated in real time.
Each step of the data processing pipeline is visualized in Fig.~\ref{fig:dije_process}.

In Algorithm \ref{alg:self_body_kmeans}, the image Jacobian of each pixel is treated as a vector (and thus the dense image Jacobian is a list of vectors), for the k-means clustering process, to create $N_\textrm{kmeans}$ cluster center vectors (i.e., \textbf{KmeansList}).
For each cluster, an evaluation value is calculated for each cluster (i.e., \textbf{EvalList}) that tries to estimate the likeliness that the cluster belongs to the robot's body.

This evaluation is updated based on an assumption that the elements of the image Jacobian are relatively consistent over time, compared to other parts of the image. The image Jacobian would still be erroneously calculated for external movement (as visible in Fig.~\ref{fig:dije_process} for the human moving in a chair in the background), but that movement does not correlate with the robot's movements, so the resulting image Jacobian rapidly changes, and the algorithm can label that part of the image as not belonging to itself.

The similarity of each cluster is calculated by first finding the closest cluster center vector from the previous result of k-means, and inherits the evaluation value from the previous corresponding cluster. The evaluation is updated based on the \textbf{Consistency} value. This uses an inverse square root (with a small value added to the denominator to avoid dividing by zero) to reward cluster centers that are closer together with the previous corresponding cluster center.
Finally, the evaluated values are normalized to $1$, and clusters with an \textbf{EvalList} value of above $e_{thresh}$ are considered to be the robot's self body ($e_{thresh} = 0.2$ was used in the experiments), and those indices are saved as \textbf{SelfBodyIndices}.

The cluster centers \textbf{KmeansList} and its indices labelled as the self-body, \textbf{SelfBodyIndices}, are used in Algorithm \ref{alg:self_body_label} to assign labels in real time. 

\begin{algorithm}
\caption{Automatic evaluation of self body from the estimated dense image Jacobian. This algorithm is run in a separate thread from the main loop.}
\label{alg:self_body_kmeans}
\begin{algorithmic}
    \Function{FindClosest}{Vec, VecList}
        \State{find i where VecList[i] is closest to Vec\\}
        \Return{i, $||$Vec - VecList[i]$||$}
    \EndFunction
    \State{Global var: KmeansList, SelfBodyIndices}
    \Let{EvalList}{$[1, 1, \dots]$}
    \While{True}
    \Let{KmeansList}{Kmeans($J^D$, $N_\textrm{kmeans}$)}
    \ForAll{Center in KmeansList}
        \Let{i, dist}{FindClosest(Center, KmeansList\_prev)}
        \Let{Eval}{EvalList[i]}
        \Let{Consistency}{$1 / \sqrt{\frac{||dist||}{||Center||} + 0.1}$}
        \Let{Eval}{Eval $\times$ $0.1($Consistency $- 1 ) + 1$}
        \State{NewEvalList.append(Eval)}
    \EndFor
    \Let{EvalList}{Normalize(NewEvalList)}
    \Let{KmeansList\_prev}{KmeansList}
    \Let{SelfBodyIndices}{Where(EvalList[i] $> e_{thresh}$)}
    \EndWhile
\end{algorithmic}
\end{algorithm}

\begin{algorithm}
\caption{Label each pixel based on the result of Algorithm \ref{alg:self_body_kmeans} in real time. Assign($\cdot$) assigns the closest cluster center to each pixel in the image (e.g. scipy.cluster.vq.vq). \textbf{SelfRecogDense} corresponds to $S_D$ in Figs.~\ref{fig:dense_jacobian_estimation} and \ref{fig:dije_process}}
\label{alg:self_body_label}
\begin{algorithmic}
    \State{Global var: KmeansList, SelfBodyIndices, SelfRecogDense}
    \While{True}
        \Let{IndexList}{Assign($[\bm j_0, \bm j_1, \dots]$, KmeansList)}
        \Let{SelfRecogDense}{IndexList is in SelfBodyIndices}
    \EndWhile
\end{algorithmic}
\end{algorithm}

\subsection{Visual servoing controller using the estimated dense image Jacobian}
\label{sec:visual_servo}
Here, we describe a markerless visual servoing controller utilizing the dense image Jacobian, that can control robots with unknown body and tool kinematic structures.
A self-body point $\bm p_{self}$ and target point $\bm p_{target}$ is defined in the image coordinates, and the controller generates joint angle update commands for the robot.
We use the letter $\bm p$ rather than $\bm x$ to describe these points, to distinguish that these are time-varying coordinates that are updated according to the actual motion, rather than the fixed coordinates of each pixel (i.e. Lagrangian vs. Eulerian frame of reference).

$\bm p_{self}$ and $\bm p_{target}$ are initially defined by the experimenter clicking on the image. $\bm p_{self}$ is updated to follow the same point on the robot, by adding the flow interpolated at that point from the dense optical flow $U_k^D$ for each timestep:
\begin{equation}
        \bm{p_{self}} \leftarrow \bm{p_{self}} + \textrm{interpolate}(U_k^D, \bm{p_{self}}
        )
\end{equation}
As long as the camera maintains a clear view of the robot body around the point $\bm p_{self}$, this was found to robustly track the same point on the robot.
Then, the image Jacobian at $\bm p_{self}$ can be calculated:
\begin{equation}
        J_{self} = \textrm{interpolate}(J_k^D, \bm{p_{self}})
\end{equation}
The pseudoinverse of the image Jacobian, $J_{self}^{\#}$, can be used to calculate a joint angle update command $\Delta{\bm q}$ that brings $\bm p_{self}$ closer to $\bm p_{target}$:
\begin{equation}
\label{eq:control-using-dije}
        \Delta{\bm q} =k_p  J_{self}^{\#} (\bm{p_{target}} - \bm{p_{self}})
\end{equation}
where $k_p$ is the feedback gain.

\section{Experiments}
\begin{figure}
    \centering
    \begin{subfigure}[b]{0.35\columnwidth}
        \includegraphics[width=\columnwidth]{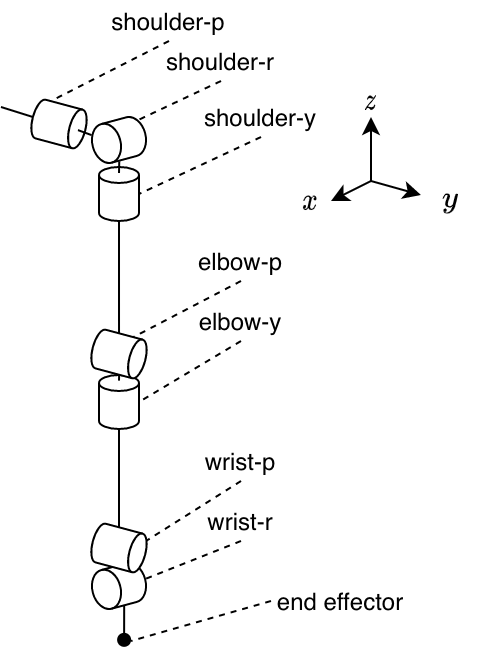}
        \caption{}
    \end{subfigure}
    \begin{subfigure}[b]{0.45\columnwidth}
        \includegraphics[width=\columnwidth]{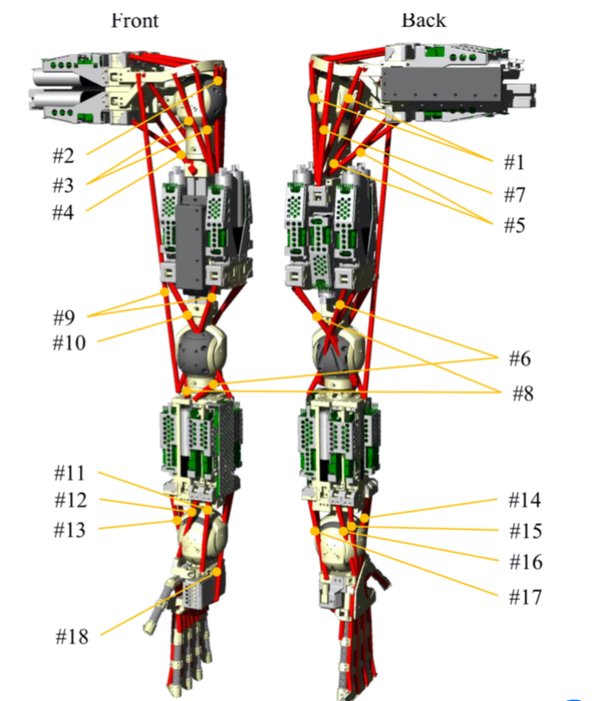}
        \caption{}
    \end{subfigure}
    \caption{Structure of musculoskeletal humanoid Musashi's arm. (a) joint and (b) muscle arrangement. \cite{Kawaharazuka2019-nt}.}
    \label{fig:muscle_joint_arrangement}
\end{figure}

The experiments were conducted on a tendon-driven musculoskeletal humanoid robot Musashi \cite{Kawaharazuka2019-nt}. Such musculoskeletal robots are actuated by flexible muscles, making them ideal for cases where the robot must adapt to holding unknown tools that cause unexpected forces \cite{Asano2016-if}. Here, controllers which map joint angle commands to muscle length commands were used as the low-level controller \cite{Kawaharazuka2018-jw, Kawaharazuka2018-ks}.
Thus, the recognition and control programs can treat Musashi as a conventional joint axis-driven robot, and we can expect that our methods work just as effectively for such axis-driven robots. The Astra S RGBD camera mounted on the head was used for input to DIJE, with a FOV of \SI{60}{\degree} $\times$ \SI{49.5}{\degree} and captures a 640 $\times$ 480 image at 30fps, and the image size was compressed in half to shorten computation time. Before each experiment, the camera angle was manually servoed to keep the point of interest in the center of the view. The depth channel was disregarded in the experiment.
The Gunnar-Farneback optical flow calculation algorithm was used to calculate the dense optical flow \cite{Farneback2003-xs}.
Fig.~\ref{fig:muscle_joint_arrangement} shows the kinematic and muscle structure of Musashi. Only some of the joints were moved in each experiment, and the other joints were controlled to keep a constant angle. DIJE and the self-recognition and visual servo control programs were implemented in Python, and could be run in real time (30fps) on a conventional notebook computer.
At the beginning of the algorithm, the value for the image Jacobian $J_0^D$ was initialized with zeros, and the covariance $\bm p_0$ was initialized with a value of $1$.
Each experiment is also shown in the video attached to this paper. 

\subsection{Robot body labeling}

\begin{figure}[tbh]
        \centering
        \includegraphics[width=0.45\textwidth]{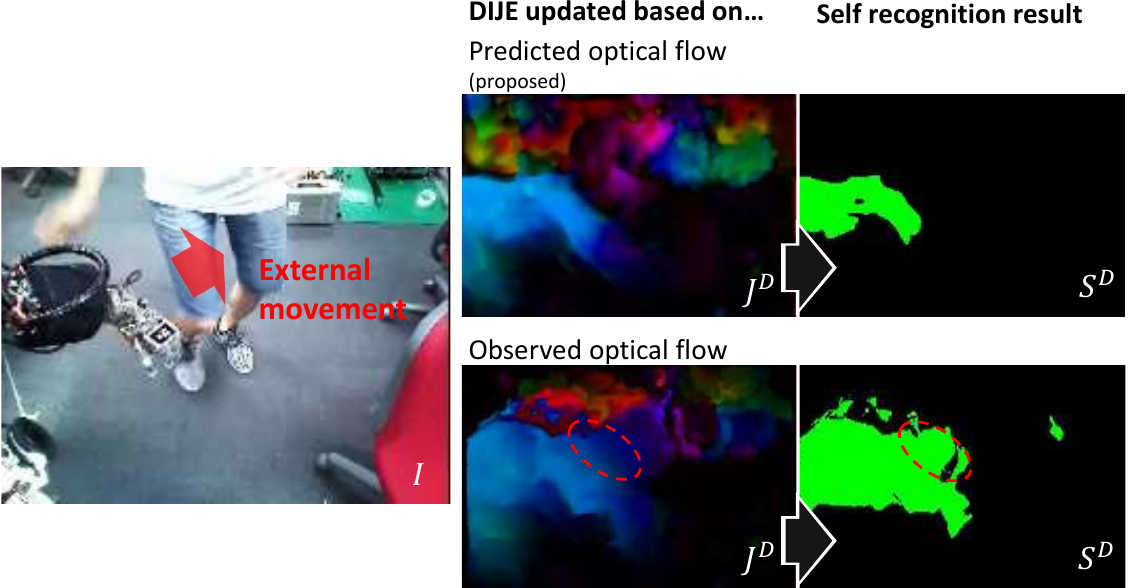}
        \caption{\label{fig:self_body_label_bleedout}Self body label "leaking out" of the actual robot body due to background movement for the observed optical flow-based update method, especially prominent in the dotted red circle area. The proposed update method does not suffer from this issue.}
\end{figure}

Here, the dense robot body labeling method introduced in Section~\ref{sec:self_body_recognition} was evaluated.
$N_\textrm{kmeans} = 5$ was used for the experiments. In theory, $N_\textrm{kmeans} = 2$ should be enough to segment the scene into the robot and the background. However, by setting a larger value for $N_\textrm{kmeans}$, the scene can be oversegmented first by the kmeans clustering algorithm and becomes more robust to various false image Jacobian values generated by background movement.

In Fig.~\ref{fig:self_body_label_bleedout}, the effect of the proposed dense image Jacobian update method used in \eqref{eq:dij_update}, using the optical flow predicted from $J^D$, was compared to the update method in \eqref{eq:dij_update_naive} which uses the measured optical flow.
It is tested in a scenario where there is overlapping movement behind the robot by a human walking.
From Fig.~\ref{fig:self_body_label_bleedout}, it can be seen that with the observed optical flow-based update method, the self-body label leaks out onto the background movement due to the optical flow caused by the human being observed outwards from the robot body. By using the proposed update method of \eqref{eq:dij_update} which uses the theoretical optical flow, the background movement is not calculated as it doesn't result from the robot's movement, and thus the algorithm can maintain a consistent label of the robot's body.

\subsection{Reaching experiments}

\begin{figure}[tbh]
\centering
\includegraphics[width=0.5\textwidth]{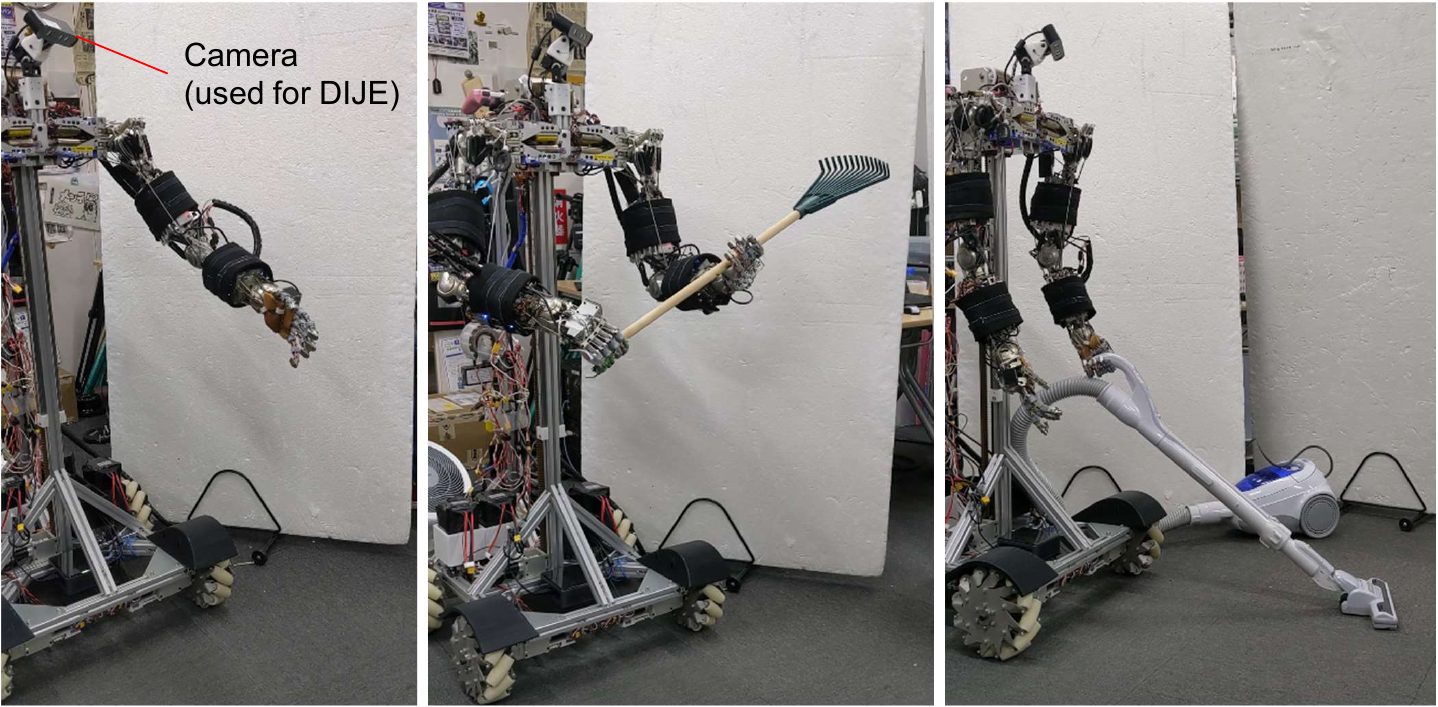}
\caption{\label{fig:dije_experiment_setup}Experiment setup for the dense image Jacobian estimation experiment. From left, tasks are: reaching, rake tip position control, vacuuming. The human experimenter placed the tools in the robot's hands in the beginning.}
\end{figure}
    
Here, the markerless visual servoing controller introduced in Section~\ref{sec:visual_servo} was run on Musashi. The feedback gain was set to $r = 0.034$, and the $\Delta{\bm q}$ joint angle update command was executed by the robot every \SI{0.7}{\second}.
\begin{figure}[tbh]
        \centering
        \includegraphics[width=0.5\textwidth]{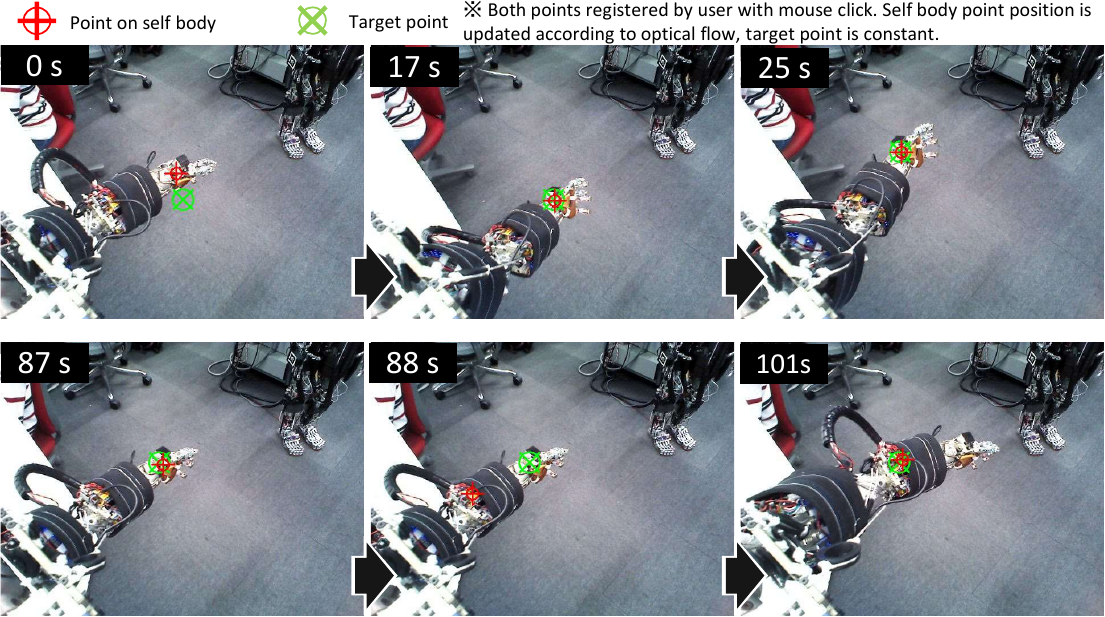}
        \caption{\label{fig:dije_result_comic}Result overlayed on camera image for the reaching experiment using DIJE-based controller. Plotted results shown in Fig.~\ref{fig:dije_result}.}
\end{figure}
In the first reaching experiment (left picture in Fig.~\ref{fig:dije_experiment_setup}), only shoulder-p, shoulder-r, shoulder-y, and elbow-p of the left arm were used.

\begin{figure}[tbh]
        \centering
        \includegraphics[width=0.45\textwidth]{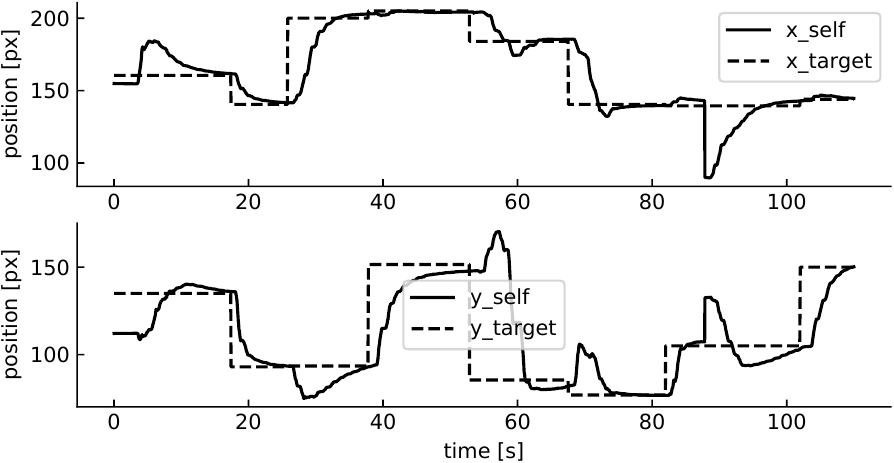}
        \caption{\label{fig:dije_result}Result of reaching experiment for the DIJE-based controller, shown in image coordinates. Visualized results shown in Fig.~\ref{fig:dije_result_comic}.}
\end{figure}

The results of the reaching experiment are shown in Fig.~\ref{fig:dije_result_comic} and \ref{fig:dije_result}. In the visualized result in Fig.~\ref{fig:dije_result_comic}, the red $+$ mark indicates $\bm p_{self}$ and the green $\times$ mark indicates $\bm p_{target}$. In the graph of Fig.~\ref{fig:dije_result}, the horizontal and vertical elements of $\bm p_{self}$ and $\bm p_{target}$ are plotted.
Right after the control is enabled at around $t = \SI{3}{s}$, $\bm p_{self}$ first moves away from $\bm p_{target}$, since the dense image Jacobian is initialized with incorrect values. However, DIJE manages to learn how to move the arm correctly, and it can follow $\bm p_{target}$ even as its position is updated. Further, at $t = \SI{88}{s}$, the position of $\bm p_{self}$ is redefined (again, by a mouse click by the experimenter) from the wrist to the elbow. Since DIJE is run densely on the image, it already knows how to move the newly defined $\bm p_{self}$, and can keep following the target.

In the next experiment, we give the robot a tool such as a rake or a vacuum cleaner, and then set $\bm p_{self}$ on the tool to control it. We used exactly the same controller as that used in the previous arm reaching experiment, except that the joints included in the dense Jacobian (and consequently the joints to be controlled) were different.
A joint angle posture that facilitates holding the tool was sent to the robot, and after the experimenter placed the tool in the robot's hands, the control process was initiated.

The robot held the rake with both hands as shown in Fig.~\ref{fig:dije_experiment_setup}. The rake is fixed to the right hand, while the left hand can freely slide along shaft of the rake. The joints shoulder-p and elbow-p of the left arm, and elbow-p of the right arm were used.
In the experiment with the vacuum cleaner, the shoulder-p, elbow-p, and wrist-p of the left arm were used.
As the head of the vacuum cleaner moves easily back and forth than in the other directions, the target points were also set along that line.

\begin{figure}[tbh]
        \centering
        \includegraphics[width=0.45\textwidth]{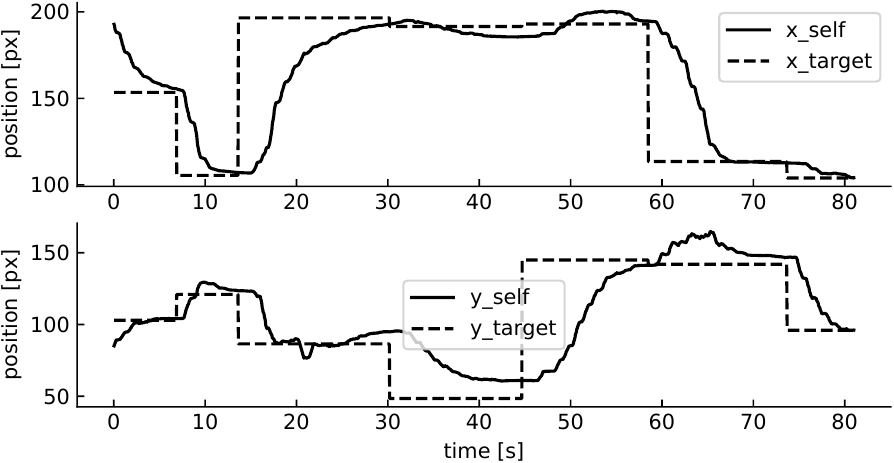}
        \caption{\label{fig:dije_rake}Result of rake tool-tip control experiment control using dense image Jacobian estimation, shown in image coordinates.}
\end{figure}

\begin{figure}[tbh]
        \centering
        \includegraphics[width=0.45\textwidth]{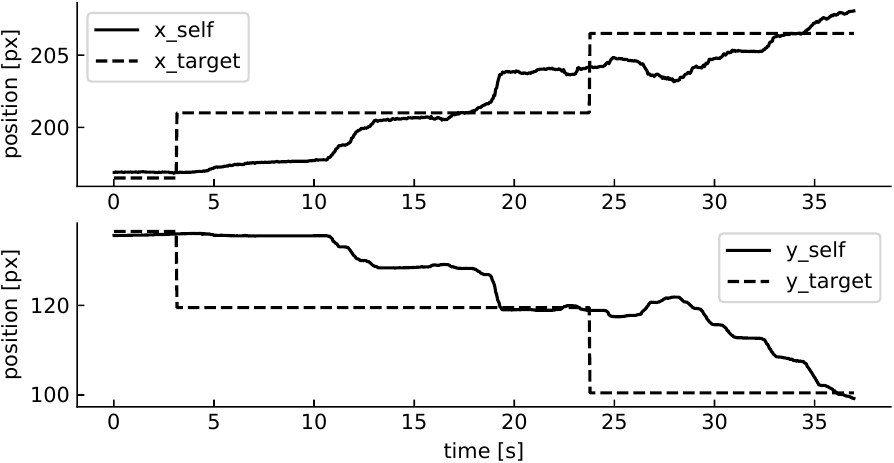}
        \caption{\label{fig:dije_vacuum}Result of vacuum tool-tip control experiment using dense image Jacobian estimation, shown in image coordinates.}
\end{figure}

The experimental results are shown in Fig.~\ref{fig:dije_rake} and Fig.~\ref{fig:dije_vacuum}.
In both experiments, we can see that $\bm p_{target}$ converges to the target point $\bm p_{target}$ after the target is updated.

\section{Discussion}
In the experiments, DIJE was able to estimate the dense image Jacobian $J^D$ appropriately and applied to the self recognition and visual servoing tasks. As can be seen in the visualization of the processing pipeline in Fig.~\ref{fig:dije_process}, when there is background movement optical flow is detected that is of similar or greater magnitude than the optical flow from the robot's movement. The DIJE process itself does not try to distinguish between robot-induced and external movement, and as visible in Figs.~\ref{fig:dije_process} and \ref{fig:self_body_label_bleedout}, the dense image Jacobian is assigned to the external movement as well. Nonetheless, since the self recognition algorithm evaluates each cluster of the dense image Jacobian based on how consistent the value is across time, it can correctly identify which part of the image belongs to the robot. We note however that this method of self recognition does not allow for recognition of non-moving parts of the robot, and it can be seen that the robot's shoulder at the lower left, which is static, is not identified as part of the robot.

The self recognition algorithm based on DIJE can also keep a consistent label even when the background movement overlaps with the robot, as seen in Fig.~\ref{fig:self_body_label_bleedout}. As previous dense self recognition algorithms were based on motion magnitude detection and did not consider the direction of motion as DIJE does, they could not distinguish when the external motion overlaps with the robot. As the representation of DIJE is higher dimension than the optical flow, it is more feature-rich and enables improved segmentation.

The markerless visual servoing controller using DIJE was also verified in an arm manipulation and tool-tip control experiment. The robot started with DIJE initialized with zero elements, and was able to quickly learn how to move the arm to the target point. Interestingly, in the rake tip control experiment we saw an emergent bimanual movement, in which by looking at only the tool tip, the robot recognizes how each joint movement relates to its movement. Although not the objective of this study, the resulting bimanual movement looks very natural and humanlike.

\section{Conclusions}
In this research, we propose the concept of a dense image Jacobian, to acquire a global representation of the robot's visuomotor coordination. It is based on the dense optical flow estimation and a simplified Kalman Filter formulation, which does not rely on markers nor \textit{a priori} knowledge of the robot's structure and can be computed in real time. The resulting high-dimensional feature has the potential to be used in a variety of recognition and control tasks, for which in this paper we propose a self-recognition algorithm and markerless visual servoing controller.

We believe that this dense visuomotor policy representation can be extended to be applied to dexterous manipulation tasks, as they can encode the relationship between the robot hand and the manipulated object in real time. One drawback of the current approach is the use of optical flow, which is purely a 2-frame image processing algorithm and does not have object constancy, making the currently method not applicable when the robot's body goes out of frame, is rotated, deformed excessively, or is occluded.
Thus, we consider combining this method with a dense visual representation which can consistently generate representations invariant to deformations \cite{O_Pinheiro2020-lz,Florence2018-jj}, to create a more general formulation of the visuomotor policy.

Further, the proposed method does not cosider depth at any point in the algorithm, and cannot distinguish errors in the depth direction. Thus, all the presented reaching experiments work in 2D, and no depth target is given. This may be resolved by calculating the optical flow in the depth direction as well, by integrating depth camera measurements or using DL-based optical flow calculations that calculate the 3D optical flow \cite{Teed2021-ha}, or by introducing an additional camera slightly offset from the first camera that can create a parallax effect encoding depth differences.
\addtolength{\textheight}{-16.8cm}   %

\bibliography{paperpile,webpages}
\bibliographystyle{IEEEtran}

\end{document}